%% file: main.tex
\documentclass[runningheads]{llncs}
\usepackage{graphicx}
\usepackage{comment}
\usepackage{amsmath,amssymb} 
\usepackage{color}

\newcommand{\R}{{\mathbb{R}}}

\usepackage{bm}
\usepackage{multirow}
\usepackage{microtype}
\usepackage{listings}
\usepackage{algorithm}
\usepackage{algpseudocode}
\usepackage{mathrsfs}

\usepackage{subfigure}
\usepackage{textcomp}
\usepackage{array,multirow,graphicx}
\usepackage{float}
\usepackage[belowskip=-3pt,aboveskip=1pt]{caption}

\begin{document}

\newcommand\mypara[1]{\vspace{0.2cm}\noindent \textbf{#1} \hspace{0.2cm}}

\pagestyle{headings}
\mainmatter
\def\ECCVSubNumber{3937}  
\newcommand{\etal}{\textit{et al.}}

\title{Self-supervised Outdoor Scene Relighting}

\begin{comment}
\titlerunning{ECCV-20 submission ID \ECCVSubNumber} 
\authorrunning{ECCV-20 submission ID \ECCVSubNumber} 
\author{Anonymous ECCV submission}
\institute{Paper ID \ECCVSubNumber}
\end{comment}

\titlerunning{Self-supervised Outdoor Scene Relighting}
\author{Ye Yu\inst{1} \and
Abhimitra Meka\inst{2} \and
Mohamed Elgharib\inst{2} \and Hans-Peter Seidel\inst{2} \and Christian Theobalt\inst{2} \and William A. P. Smith\inst{1}}
\authorrunning{Y. Yu et al.}

\institute{University of York, United Kingdom\\
\email{\{yy1571,william.smith\}@york.ac.uk} \and
Max Planck Institute for Informatics, Saarland Informatics Campus, Germany \\
}
\maketitle

\begin{abstract}
Outdoor scene relighting is a challenging problem that requires good understanding of the scene geometry, illumination and albedo. Current techniques are completely supervised, requiring high quality synthetic renderings to train a solution. Such renderings are synthesized using priors learned from limited data. In contrast, we propose a self-supervised approach for relighting. Our approach is trained only on corpora of images collected from the internet without any user-supervision. This virtually endless source of training data allows training a general relighting solution. Our approach first decomposes an image into its albedo, geometry and illumination. A novel relighting is then produced by modifying the illumination parameters. Our solution capture shadow using a dedicated shadow prediction map, and does not rely on accurate geometry estimation. We evaluate our technique subjectively and objectively using a new dataset with ground-truth relighting. Results show the ability of our technique to produce photo-realistic and physically plausible results, that generalizes to unseen scenes. 
\end{abstract}

\section{Introduction}
\label{sec:intro}
\input{1_introduction.tex}

\section{Related Work}
\label{sec:related}
\input{2_related_work.tex}

\section{Overview}
\label{sec:overview}
\input{3_overview.tex}

\input{4_technical_section.tex}

\section{Results}
\label{sec:results}

\input{5_results.tex}

\section{Discussion}
\label{sec:discussion}
\input{7_discussion.tex}

\section{Conclusion}
\label{sec:conclusion}
\input{8_conclusion.tex}

\section*{Acknowledgments}
This work was funded by the ERC Consolidator Grant \textit{4DRepLy} (770784). 

\bibliographystyle{splncs04}
\bibliography{refs}
\end{document}

%% file: 1_introduction.tex
Virtual relighting of real world outdoor scenes is an important problem that has wide applications. Performing such a relighting task involves correctly estimating and editing the various scene components -- geometry, reflectance and the direct and indirect lighting effects. Measuring these high-dimensional parameters traditionally required the use of  instruments such as LIDAR scanners and gonio-reflectometers and extensive manual effort \cite{Tchou:2004:UP:1186223.1186323,Troccoli2008}. This problem has been simplified by using only a small number of 2D images of a scene in a process known as image based rendering (IBR), but this leads to far fewer constraints on the unknown variables and runs into the problem of ill-posedness. 

\begin{figure}[!t]
    \centering
        \begingroup
\setlength{\tabcolsep}{1pt}
\renewcommand{\arraystretch}{0.5}
\resizebox{\textwidth}{!}{
\begin{tabular}{ccccccccc}
\includegraphics[width=1.2cm]{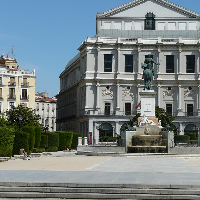}&
\includegraphics[width=1.2cm]{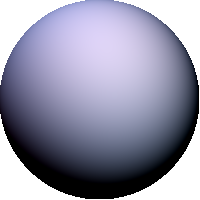}&
\includegraphics[width=1.2cm]{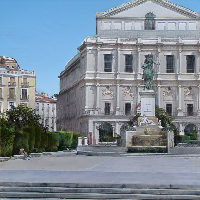}&
\includegraphics[width=1.2cm]{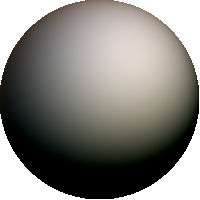}&
\includegraphics[width=1.2cm]{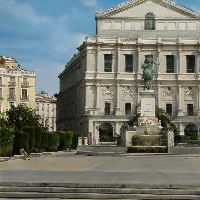}&
\includegraphics[width=1.2cm]{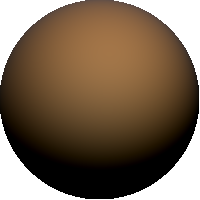}&
\includegraphics[width=1.2cm]{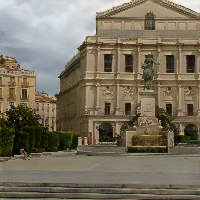}&
\includegraphics[width=1.2cm]{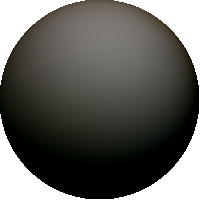}&
\includegraphics[width=1.2cm]{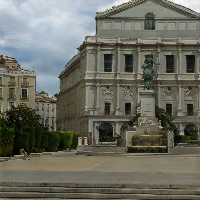}\\
\includegraphics[width=1.2cm]{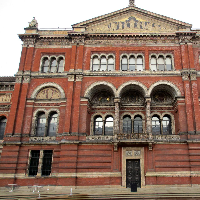}&
\includegraphics[width=1.2cm]{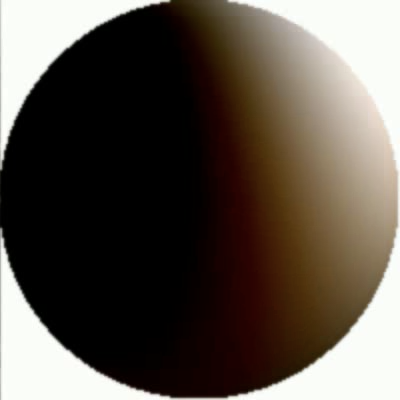}&
\includegraphics[width=1.2cm]{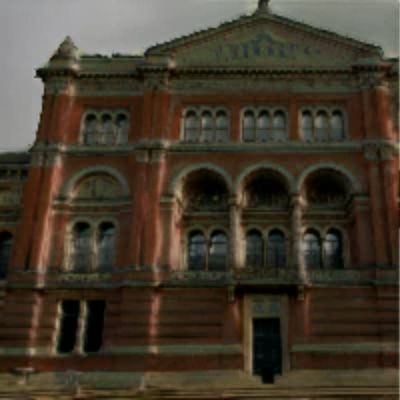}&
\includegraphics[width=1.2cm]{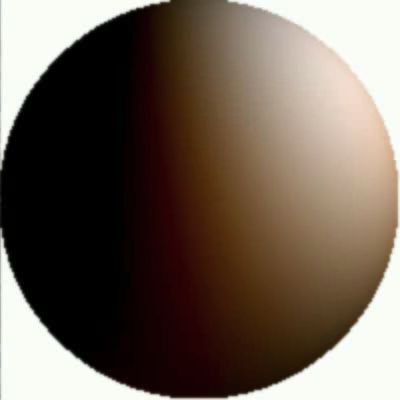}&
\includegraphics[width=1.2cm]{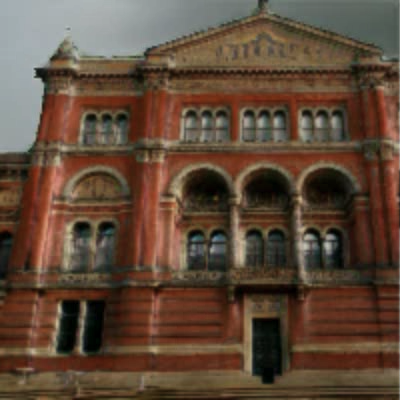}&
\includegraphics[width=1.2cm]{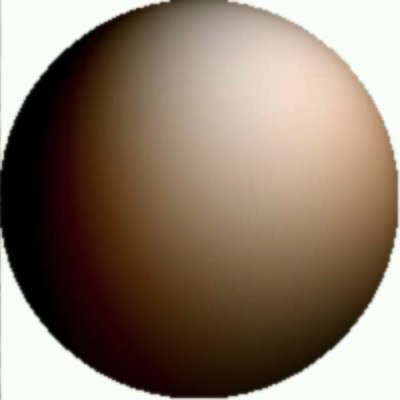}&
\includegraphics[width=1.2cm]{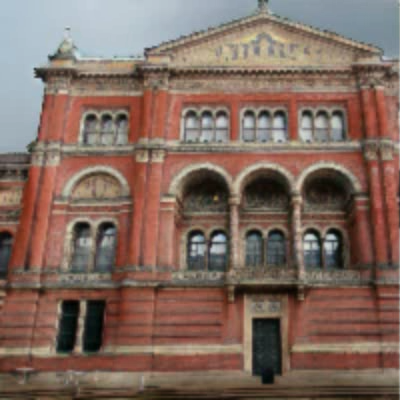}&
\includegraphics[width=1.2cm]{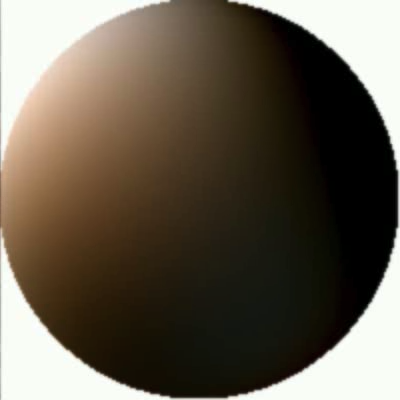}&
\includegraphics[width=1.2cm]{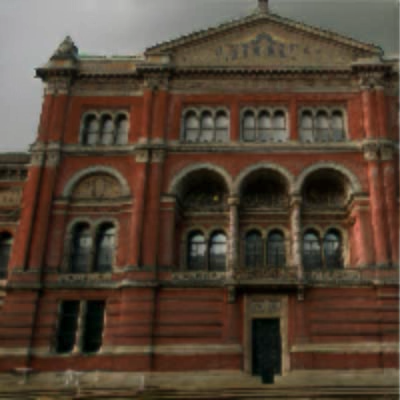}\\
{\tiny Input} & {\tiny Illum.1} & {\tiny Relighting1} & {\tiny Illum.2} & {\tiny Relighting2} & {\tiny Illum.3} & {\tiny Relighting3} & {\tiny Illum.4} & {\tiny Relighting4} \\
\end{tabular}
}
\endgroup
    \caption{We present a novel self-supervised technique to photorealistically relight an outdoor scene from a single image to any given target illumination condition. Our method is able to generate plausible shading, shadows, color-cast and sky region in the output image, while preserving the high-frequency details of the scene reflectance.}
    \label{fig:teaser}
\end{figure}

Multi-view and multi-illumination constraints have proved to be effective in solving this problem  \cite{Laffont:2012:CII:2366145.2366221,Duchene:2015:MII:2843519.2756549,Philip:2019:MRU:3306346.3323013,Yu_2019_CVPR}. 2D images of a scene from different viewpoints and under different lighting conditions provide the necessary constraints to reconstruct the geometry of the scene and disambiguate the lighting from the reflectance. For example, the method of Laffont \etal~\cite{Laffont:2012:CII:2366145.2366221}, along with multi-view 3D reconstruction, also uses manual interactions to perform an intrinsic decomposition of the scene images into reflectance and shading layers. By reprojecting the reflectance layer from one viewpoint to another and recombining with the original shading image, lighting conditions of one image of a scene can be transferred to another. While this technique is effective, it is limited in its relighting capability because it cannot relight the scene under an arbitrary lighting condition of choice. The method of Duchene \etal~ \cite{Duchene:2015:MII:2843519.2756549} also performs a similar intrinsic decomposition of multi-view images, and additionally estimate the shadows and the parameters of a sun-lighting model for the scene. These parameters are then modified in a geometrically accurate way to achieve scene relighting. Philip \etal~ \cite{Philip:2019:MRU:3306346.3323013} similarly estimate shadows and sun-light model parameters, but skip the inverse rendering process and instead use a deep neural network to directly generate relighting results. Their network takes as input several `illumination buffers' that are rendered using the reconstructed geometry and estimated sun-light model parameters. This method relies on high-quality ground-truth renderings of synthetic 3D models of outdoor scenes, requiring the availability of high-end computational hardware. While these techniques have been shown to generate high-quality relighting results on real scenes, they are limited by the availability of a multi-view images of the scene. They also rely on a sun-lighting model that only works for bright sunlight conditions and does not generalize to cloudy overcast skies, night-time lighting, or other desired target illumination conditions. 

Another class of methods circumvent the problem of estimating the scene parameters by achieving relighting directly through lighting style transfer. These methods \cite{Shih:2013:DHD:2508363.2508419,Lalonde:2009:WCA:1661412.1618477} change the lighting in a scene by learning the colour characteristics of images at different times of the day. Another set of methods \cite{Kopf:2008:DPM:1409060.1409069,8100223} learn a more general class of style-transfer in which characteristics of a reference image are transferred to a target image, including the scene lighting. Such methods are not physically based and are limited in relighting a scene either based on a reference image or a particular time of the day.

In contrast, the method of Yu and Smith \cite{Yu_2019_CVPR} proposes a novel formulation for the problem that allows for fully controlled relighting based on a single image of the scene. They demonstrate a learning method that at training time uses the constraints available from multi-view casual images of outdoor scenes sourced from the internet, to learn to estimate the scene appearance parameters. The network can then at test time estimate these parameters from a single image. By modifying the lighting to a desired lighting environment, the image can be relit. While this method enabled relighting of a scene from a single 2D image to any arbitrary lighting, it was also limited by the low-frequency lighting model used in the decomposition that lead to non-photorealistic relighting results.

Recently, the advent of adversarial learning technique \cite{NIPS2014_5423} has enabled neural networks to generate photorealistic images. `Neural rendering' techniques based on this principle have shown promising results in various allied tasks such as novel-view synthesis \cite{Martin-Brualla:2018:LEP:3272127.3275099}, view-dependent effects rendering \cite{thies2019neural} and appearance modification \cite{48577}.

Motivated by these two advances, we propose the first fully self-supervised neural rendering framework for performing photorealistic relighting of an outdoor scene from a single image with full lighting controllability (see Figure \ref{fig:teaser}). Similar to the method of Yu and Smith \cite{Yu_2019_CVPR}, our method learns to estimate scene appearance parameters based on multi-view constraints at training time, without using any ground-truth synthetic 3D renderings. At test time, it takes as input a single 2D image and estimates the underlying appearance parameters such albedo, shading, shadows, lighting and normals. These physical parameters are then fed to a novel neural rendering framework, along with target lighting conditions, to generate photorealistic relighting of the scene and the sky region. By training our system in a completely self-supervised manner, it generalizes to unseen novel scenes and any target lighting condition of choice as provided by the user in the form of an environmental light map. We introduce a new high-resolution HDR multi-view \& multi-illuminant evaluation dataset for outdoor relighting, and our extensive test results on the dataset show the efficacy of our method.
 
In summary, our main technical contributions are:
\begin{itemize}
	\item The first fully-automatic single-image based relighting technique for outdoor scenes with full controllability of target lighting
	\item A novel self-supervised neural rendering framework that uses physical intrinsic decomposition layers of the scene to generate photorealistic relighting results without using any ground-truth data or synthetic 3D rendering 
	\item A sky generation network that generates plausible sky region for the scene under a given target lighting environment
	\item A high-quality evaluation dataset for outdoor relighting with ground-truth HDR environment maps.
\end{itemize}

%% file: 2_related_work.tex
Relighting a scene is a complex task. In order to perform physically accurate relighting, all components of light-transport in the scene need to be measured and modified, in a process known as inverse rendering \cite{doi:10.1111/j.1467-8659.2003.00716.x}. Traditionally, this involved using special optical equipment to measure the geometry \cite{Yu:1998:RPP:280814.280874,loscos1999interactive,meka:2017}, surface reflectance \cite{Debevec:2000,Wenger05,Masselus:2003:RIL:882262.882315,Troccoli2008,Meka19} and environmental illumination \cite{Debevec:2002:IL:616075.618886,HoldGSHGL2017,Lalonde2012,Stumpfel:2006:DHC:1185657.1185687}, while also inverting the global illumination within the scene \cite{YuDMH1999}. Image-based relighting techniques have attempted to simplify the problem by using only 2D images for the task. But using only 2D images makes the problem highly under-constrained and ambiguous.

Due to the ambiguous nature of the problem, recently there has been a lot of interest in applying learning based methods to solving it \cite{DPR,shu2017neural,Garon_2019_CVPR,8953635}. We restrict our discussion to methods that perform scene level relighting.   
Due to the very different nature of geometry and illumination in indoor and outdoor scenes, the two have often been treated as separate class of inverse rendering problems. Inverse Rendering in outdoor scenes has usually dealt with specific illumination models for natural illumination. \cite{doi:10.1111/cgf.12217} propose a single-image approach that accounts for environment lighting in outdoor scenes. Collections of photographs of a scene have been used to provide better constraints for relighting \cite{5206753,Shan:2013:VTT:2544744.2544787}. While we also use a dataset of casual photography of particular scenes to learn to perform inverse rendering and relighting, but at test time, we only rely on a single image of a scene to perform photorealistic relighting. The method of \cite{Shih:2013:DHD:2508363.2508419} performs lighting transfer by matching a single image to a large database of timelapses, but cannot treat cast shadows.
Alternatively, online digital terrain and urban models registered to images can be used for approximate relighting \cite{Kopf:2008:DPM:1409060.1409069}. 

Several methods on multi-view image relighting have been developed, both for the case of multiple images sharing single lighting
conditions \cite{Duchene:2015:MII:2843519.2756549}, and for images of the same location with multiple lighting conditions (typically from internet
photo collections) \cite{Laffont:2012:CII:2366145.2366221}. For the single lighting condition, \cite{Duchene:2015:MII:2843519.2756549}, first perform shadow
classification and intrinsic decomposition using separate optimization steps. Despite impressive results, artifacts remain especially
around shadow boundaries and the relighting method fails beyond
limited shadow motion. More recently, several learning based methods have been suggested to perform relighting in outdoor scenarios \cite{Yu_2019_CVPR,Xu:2018:DIR:3197517.3201313,Philip:2019:MRU:3306346.3323013,DBLP:journals/corr/abs-1901-02453}.
A simpler version of the relighting problem, is of integrating virtual objects into real scenes in an illumination-consistent manner, have been solved by using proxy geometry and user interaction \cite{Karsch:2011:RSO:2024156.2024191,doi:10.1111/cgf.12217,Okabe06single-viewrelighting,Meka:2018}. But these methods do not solve the problem of general relighting of scenes. Webcam sequences have also been used for relighting \cite{Lalonde:2009:WCA:1661412.1618477,Sunkavalli:2007:FTV:1275808.1276504}, although cast shadows often require manual layering.

%% file: 3_overview.tex
\begin{figure}[!t]
    \centering
    \includegraphics[width=0.9\textwidth,clip=true,trim=15px 335px 40px 120px]{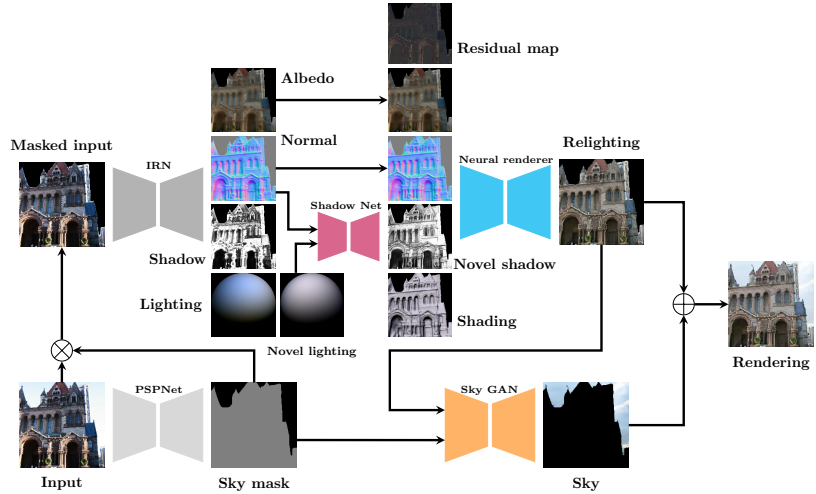}
    \caption{For a given \textit{Input} image of an outdoor scene, our method first performs a physical decomposition of the scene into various components. Using a pre-trained segmentation network (\textbf{PSPNet} \cite{zhao2017pyramid}), the scene is separated from the sky. The scene is then decomposed by the \textbf{InverseRenderNet (IRN) \cite{Yu_2019_CVPR}} into intrinsic image layers of \textit{Albedo}, \textit{Normal}, \textit{Shadow} and \textit{Lighting} . Given a target \textit{Novel lighting} condition, \textbf{ShadowNet} uses the regressed scene normals to generate a target \textit{Novel shadow} map for the scene. The scene albedo and normals, along with target lighting, shadow map, target shading and residual input map (see Section \ref{sec:neural}) are then fed to the \textbf{Neural renderer} to generate plausible \textit{Relighting} of the scene. Given the output of the neural renderer, \textbf{SkyGAN} generates a convincing \textit{Sky} region, and by compositing these together, a complete photorealistically relit \textit{Rendering} is achieved.}
    \label{fig:overview}
\end{figure}

Neural inverse rendering has been recently shown to enable convincing decomposition of both indoor \cite{sengupta2019neural} and outdoor \cite{Yu_2019_CVPR} uncontrolled scenes into geometry (normal map), illumination and reflectance. These methods are self-supervised via a physics-based model of image formation. Such models are typically based on simple assumptions such as perfect Lambertian reflectance and ignore global illumination effects and shadowing. For this reason, re-illumination of the geometry and reflectance with novel lighting does not lead to photorealistic images. In addition, sky regions do not adhere to reflectance models, and so they are either missing from relighting results or the original sky is pasted back, making it inconsistent with the new lighting.

Our goal in this paper, motivated by the recent advances in neural inverse rendering \cite{Yu_2019_CVPR}, is to learn in a fully self-supervised fashion to perform photorealistic relighting of outdoor scenes from a single image. Photorealism is achieved by replacing classical model-based renderers with a learnt neural renderer that can take as input the various scene parameters along with a target lighting condition and generate plausible relighting results. The neural renderer particularly learns to synthesize global illumination effects such as plausible shadows, inter-reflections and view-dependent effects that are required for photorealism, which are much more difficult to simulate with model-based renderers. The neural renderer is trained using an adversarial loss to ensure that the generated images lie within the distribution of real images. A novel cycle consistency loss and direct supervision loss via cross projection of multi-view images is also used to ensure that the generated images exhibit the desired target lighting. We also present a sky generation network that learns to synthesize plausible skies that are consistent with the lighting within the rest of the image. An overview of our approach is shown in Figure \ref{fig:overview}.

%% file: 4_technical_section.tex
\section{Inverse rendering}\label{sec:IRN}

We take as our starting point the inverse rendering network of Yu and Smith \cite{Yu_2019_CVPR}. InverseRenderNet comprises an image-to-image network that estimates colour diffuse albedo, ${\bm \alpha}(p)=[\alpha_r(p), \alpha_g(p), \alpha_b(p)]^T$, and surface normal direction, $\mathbf{n}(p)\in\R^3$, $\|\mathbf{n}(p)\|=1$, for each pixel $p$. Illumination is represented using the parameters, $\mathbf{L}\in\R^{3\times 9}$, of an order 2 spherical harmonics model \cite{Ramamoorthi:2001} leading to the following image formation model:
\begin{equation}
    \mathbf{i}(p) = {\bm \alpha}(p) \odot \mathbf{L}\mathbf{b}(\mathbf{n}(p)),\label{eqn:IRNmodel}
\end{equation}
where $\mathbf{b}(\mathbf{n}(p))\in\R^9$ contains the spherical harmonic basis for normal direction $\mathbf{n}(p)$, $\odot$ is the elementwise product and $\mathbf{i}(p)$ the RGB colour at pixel $p$ 
$\mathbf{L}$ is computed by solving a least squares system over all foreground pixels, $p\in\mathcal{F}$, i.e.~those not labelled as sky by a PSPNet segmentation network \cite{zhao2017pyramid}. $\mathbf{L}$ is further restricted to a statistical subspace learnt from real, outdoor environment maps. The self-supervision loss is provided by the residual error in \eqref{eqn:IRNmodel}.

In the context of relighting, the main drawback of InverseRenderNet is that the model used for self-supervision cannot adequately describe real world appearance. So, unmodelled phenomena such as cast shadows, spatially varying illumination and specularities are baked into
albedo and normal maps. Of these phenomena, the most severe are shadows. When baked into the albedo map, relit images retain the shadows of the original illumination. When baked into the normal map, relit images contain shading artefacts caused by warped normals. 

We propose a novel variant of InverseRenderNet that explicitly estimates an additional channel, $s(p)$, to explain these unmodelled phenomena and avoid them being baked into the albedo or normal maps. The additional channel acts multiplicatively on the appearance predicted by the local spherical harmonics model:
%\
\begin{equation}
    \mathbf{i}(p) = s(p){\bm \alpha}(p) \odot \mathbf{L}\mathbf{b}(\mathbf{n}(p)).
\end{equation}
Without appropriate constraint, the introduction of this additional channel could lead to trivial solutions. Hence, we constrain it in two ways. First, we restrict it to the range $[0,1]$ so that it can only downscale appearance. Second, it is a scalar quantity acting equally on all colour channels. Together, these restrictions encourage this channel to explain cast shadows and we refer to it as a shadow map. However, note that we do not expect it to be a physically valid shadow map nor that it contains only shadows. During training, we compute our self-supervised appearance loss in a shadow free space:
\begin{equation}
    \ell_{\text{appearance}} = \sum_p \left\| \min\left(1,\frac{\mathbf{i}(p)}{s(p)}\right) - {\bm \alpha}(p) \odot \mathbf{L}\mathbf{b}(\mathbf{n}(p)) \right\|^2,\label{eqn:appLoss}
\end{equation}
i.e.~we compare the appearance predicted by the local illumination model against the original image with shadows divided out.

\begin{figure}[!t]
    \centering
\begingroup
\setlength{\tabcolsep}{1pt}
\renewcommand{\arraystretch}{0.5}
\small{
\resizebox{\textwidth}{!}{
\begin{tabular}{cccccccccc}
Input & albedo & Normal & Shadow & \footnotesize{Shadow free} & Input & albedo & Normal & Shadow & Shadow free  \\
\includegraphics[width=1.6cm]{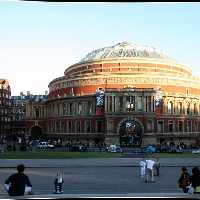}&
\includegraphics[width=1.6cm]{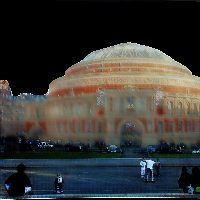}&
\includegraphics[width=1.6cm]{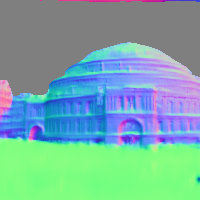}&
\includegraphics[width=1.6cm]{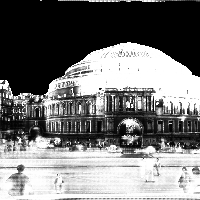}&
\includegraphics[width=1.6cm]{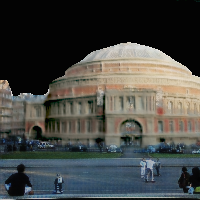}&
\includegraphics[width=1.6cm]{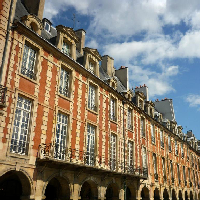}&
\includegraphics[width=1.6cm]{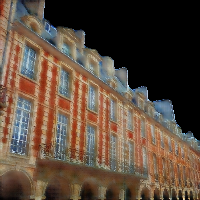}&
\includegraphics[width=1.6cm]{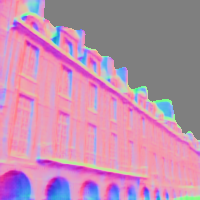}&
\includegraphics[width=1.6cm]{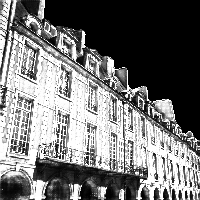}&
\includegraphics[width=1.6cm]{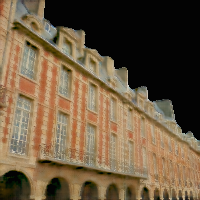}\\
\includegraphics[width=1.6cm]{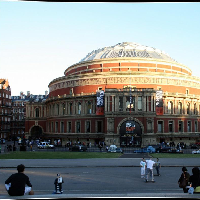}&
\includegraphics[width=1.6cm]{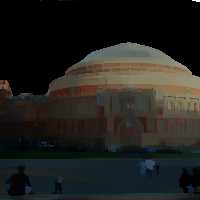}&
\includegraphics[width=1.6cm]{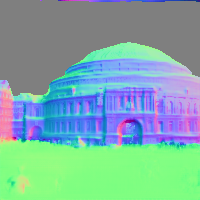}&
&
\includegraphics[width=1.6cm]{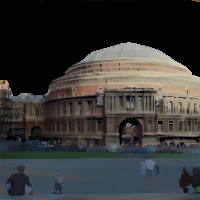}&
\includegraphics[width=1.6cm]{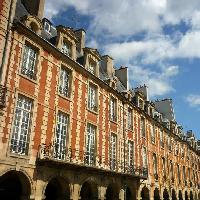}&
\includegraphics[width=1.6cm]{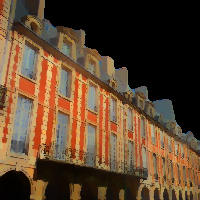}&
\includegraphics[width=1.6cm]{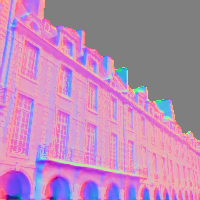}&
&
\includegraphics[width=1.6cm]{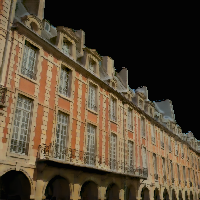}\\
&
\end{tabular}
}}
\endgroup
    \caption{Inverse rendering with shadow prediction. Rows 1: proposed variant, rows 2: original InverseRenderNet \cite{Yu_2019_CVPR}.}
    \label{fig:IRNshadows}
\end{figure}

For training, we use the same dataset, training schedule and hyperparameters as the InverseRenderNet and retain multiview supervision. Specifically, albedo consistency and direct normal map supervision are applied in the same way while the cross-rendering loss (mixing lighting from one view with albedo and normal map from another) is formulated in the shadow free space as in \eqref{eqn:appLoss}. 
Explicitly modelling shadow allows us to drop generic priors used in InverseRenderNet (albedo smoothness and pseudo-supervision) such that addressing problems like oversmoothing.
We show some sample qualitative results in Figure \ref{fig:IRNshadows}. Note that, relative to InverseRenderNet, cast shadows are not baked into the albedo map such that the shadow free rendering removes their effect.

\section{Neural rendering}
\label{sec:neural}

We now describe our neural rendering network. This can be viewed as a conditional GAN \cite{mirza2014conditional} in which the conditioning input is the maps required for a Lambertian rendering and the latent space is the spherical harmonic lighting parameter space. The objective of the network is to generate images indistinguishable from real ones while keeping the lighting consistent with the target lighting parameters.

The input to the neural rendering network is constructed from the outputs of InverseRenderNet (see Figure. \ref{fig:overview}). The albedo and normals are taken as direct inputs from the output of InverseRendernet, because they are scene invariants. Additional inputs of a shading map and a shadow map consistent with the target illumination are constructed. The shading channel is obtained using the Lambertian spherical harmonic lighting model under the desired lighting with the estimated normal map. The shadow map for a given novel lighting condition is predicted using a separate shadow prediction network described in Section \ref{sec:shadownet}.

We concatenate the albedo prediction (3 channels), normal prediction (3 channels), shading (3 channels), shadow map (1 channel) and sky segmentation (1 channel) into an 11 dimensional tensor. In addition to this tensor, we compute another 3-channel \textit{residual map} that contains the lost fine-scale details from original image after inverse rendering decomposition. The residual map is computed by subtracting Lambertian rendering composed by inverse rendering results from original input image. We then stack this residual map at the end of concatenated 11 dimensional tensor and feed it to the neural rendering network.

\subsection{Losses} We use three classes of loss function in order to train the neural renderer. First, an adversarial loss ensures the realism of the generated images. Second, direct supervision is provided in the form of self-reconstruction and cross-projection rendering losses to ensure the images are accurate predictions of the scene appearance under desired lighting conditions. Third, this direct supervision is aided by a cycle consistency loss that uses InverseRenderNet to consistent decompositions of original and rendered images.

\mypara{Adversarial loss} For adversarial loss we use the multiscale LSGAN \cite{mao2017least} architecture. Real images are true images with the sky masked out. Fake images are the neural renderings, again with all pixels in the sky region set to black. 

\mypara{Direct supervision} Our training set provides real example images under a variety of illumination conditions. We can exploit these for direct supervision. When the chosen lighting condition for relighting is the same as the original image, we expect the neural rendering to exactly match the original image. We refer to this as self-reconstruction loss. In practice, this is computed as a sum of the VGG perceptual loss \cite{Simonyan15} (difference in VGG features from the first two layers) and $\ell_{2}$ distance in LAB colour space.
However, self-reconstruction loss does not penalise baked-in effects. To overcome this, we use multiview supervision. A mini-batch consists of a set of overlapping images with different illumination and which can be cross projected from one view to another using the multi-view stereo (MVS) reconstructed geometry and camera parameters. We use this for additional direct supervision. Within a mini-batch, we shuffle the lighting estimates from InverseRenderNet so that we relight the albedo and normal predictions from one view with the lighting from another. We rotate the spherical harmonic lighting to account for the relative pose between views. Supervision is provided by comparing the neural rendering against the cross projection of the view from which the lighting was taken, again measured in terms of VGG perceptual loss and $\ell_{2}$ distance in LAB space. However, errors in the MVS geometry and camera poses cause slight misalignments in the cross projected images. We found that applying this loss at full resolution led to a blurry output. For this reason, before computing the cross projection loss, we downscale both the cross projected and rendered images by a factor of $4$.

\mypara{Cycle consistency} We found that direct supervision and adversarial loss alone are insufficient for good performance and smooth relighting under smooth illumination parameter changes. This is partly due to the fact that cross projected images are incomplete and can be quite sparse when the view change is large. Therefore, to improve stability we propose to also include a cycle consistency loss. Here, we use the InverseRenderNet trained as described in Section \ref{sec:IRN} and measure the consistency between the input maps to the neural renderer and those obtained by decomposing the neural rendered image. Specifically, we penalise the difference in the albedo, normal, lighting and shadow maps. Lighting consistency is measured by the sum of VGG perceptual loss and $\ell_{2}$ difference between the Lambertian shading maps. Normal map consistency is measured by the mean angular error between original and estimated normal maps. For albedo consistency, we weight the error by the shading map. The idea is that albedo estimates in darkly shaded regions are unlikely to be accurate and we do not wish to overemphasise errors in these regions. Again, the albedo difference is measured in terms of VGG perceptual loss and $\ell_{2}$ distance in LAB space.

\subsection{Shadow prediction network}
\label{sec:shadownet}

When illumination changes, the shadowing changes. 
To estimate such changes in shadows, we train a separate shadow prediction network. It takes as input a normal map and the spherical harmonic lighting vector and outputs a shadow map. In order to input the lighting vector while retaining the image-to-image architecture of the network, we replicate the 27D lighting vector (since $\mathbf{L}\in\R^{3\times 9}$) pixel-wise and attach it to normal map such that the input is a 30D tensor. We train the shadow prediction network using illumination, normal and shadow maps predicted by our modified InverseRenderNet.

\subsection{Sky GAN}
\label{sec:skygan}

Our physical illumination model is only able to describe non-sky regions of the image. Sky cannot be meaningfully represented in terms of geometry, reflectance and lighting. Moreover, sky appearance is partially stochastic (the precise arrangement of clouds is not informative). 
For this reason, we train a second network specifically to generate skies that are plausible given the rest of the image. For example, if the image contains strong cast shadows and shading, one would expect a clear sky with sunlight coming from an appropriate direction. If the image is highly diffuse with little discernible shading one would expect an cloudy sky.

For this purpose, we use the GauGAN architecture \cite{park2019SPADE} with two semantic classes: sky and foreground. This network performs sky generation from random noise and conditional inputs of the sky segmentation mask and the foreground image with black sky. The output is the sky image which is blended with the foreground image using the binary sky mask. Such binary blended images are inputs to the discriminator along with the sky mask as a conditional input. Hence, the discriminator loss will help generate both more realistic skies but also skies that are plausible given the foreground appearance.

To train the generator, we use the adversarial loss and the feature matching loss as in \cite{park2019SPADE} but remove other appearance losses. We train using real images in which sky has been masked to black. The discriminator is trained using the same loss as the original GauGAN \cite{park2019SPADE}. We find that, in practice, this network generalises well to foregrounds generated using our neural rendering network.

\subsection{Training}

Our network graph is implemented in tensorflow. The neural rendering network and shadow prediction network are modelled as UNET architectures \cite{RonneFB2015} and The skyGAN network and InverseRenderNet were modelled after  ResNet architecture \cite{DBLP:journals/corr/HeZRS15}. For details of our network architectures and training hyperparameters, please refer to the supplementary document.

The training of the networks is performed in several stage. The inverse rendering network is trained indepedently as the first step. The output of inverse rendering network is used to train the shadow prediction network. Given the well-trained shadow prediction network and inverse rendering network, the neural rendering network is trained. The training of the neural rendering network is done in two phases. In the first phase only a self-reconstruction loss is employed and this stage is stopped when the loss reaches a steady-state value. In the second phase, the cycle-consistency loss and adversarial loss are added. In the experiments, we found such pre-training step ensures fast convergence and leads to renderings containing more fine details.

Similar to Yu and Smith \cite{Yu_2019_CVPR}, we run our training and testing on the megaDepth dataset \cite{MegaDepthLi18}. The dataset contains multiview stereo images, which enable us to directly train inverse rendering network and find relative rotations between image views before shuffling illumination estimates. The dataset contains a variety of  outdoor scenes. All training images were resized to a size of $200\times200$ pixels to keep the training tractable on single-gpu hardware.

%% file: 5_results.tex
\subsection{Outdoor Relighting Bechmarking Dataset}
\label{sec:dataset}

\begin{figure}[!t]
	\centering
	\includegraphics*[width=0.8\linewidth]{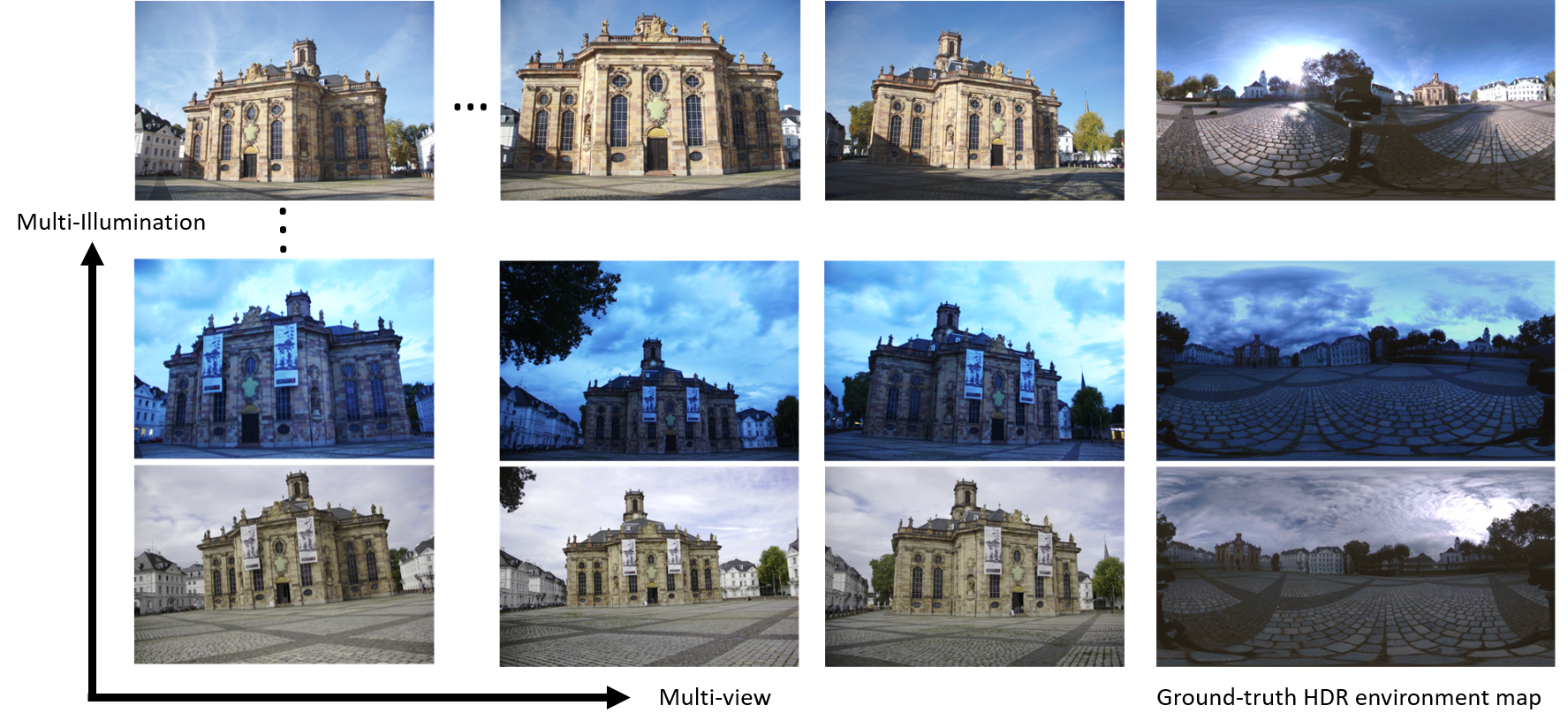}
	\caption{We present a new high-quality high-resolution outdoor relighting dataset. Our dataset consists of high-resolution HDR images of a single monument captured under several different lighting conditions from multiple views, along with the ground-truth HDR environment light maps.}
	\label{fig:dataset}
\end{figure}

We present a new high-quality benchmarking dataset for the evaluation of outdoor relighting techniques. The dataset consists of several sets of multi-view, multi-illumination high dynamic range (HDR) images of a single monument, along with ground-truth HDR environment maps for each illumination condition. 
We captured 6 different lighting conditions, including clear sky with bright sunlight, cloudy overcast sky and evening light.  For each lighting condition, we capture 10 images from views around the monuments and also the ground-truth environment light map. Each image of the monument is of resolution $5184 \times 3456$, captured with a Canon 5D Mark II DSLR camera with an 18mm focal length lens. It consists of 6 multi-exposure raw captures, which are fused in Adobe Photoshop to generate an HDR image. The lowest camera exposure time is chosen to ensure that the captured image has minimal amount of pixel saturation from bright light sources such as the sun. We use constant ISO and aperture settings in the capture. The environment light map is captured using a 360 degree camera (LG360) with 6 multi-exposure shots fused to obtain the HDR image. 

While the original environment maps are captured from arbitrary viewpoints, in order to perform view consistent relighting, the environment maps need to be rotated to align them to the same viewpoint as the camera images. This is achieved by  performing multi-view 3D reconstruction of the monument from all the dataset images and estimating accurate camera pose for each camera view through bundle adjustment. The rotation between environment map and the global co-ordinate system of the monument (taken as the camera co-ordinate system of the first camera view image) is computed by performing a sparse feature match between the environment map and the 3D model and optimizing for the camera rotation between the two. This process is repeated for each of the 6 lighting conditions. In the dataset, we provide the camera pose for every image and also the rotation for each of the 6 ground-truth environment maps to the first camera view image. This provides `aligned environment maps' for each lighting condition. Please see the supplementary document for an illustration of this alignment process. A low-frequency representation of each captured environment map is also provided by computing the 2nd order spherical harmonics co-efficients that fit the light map.

\subsection{Qualitative Evaluation}

\begin{figure}[!t]
	\centering
	\includegraphics[width=\linewidth,clip=true,trim=10px 112px 210px 100px]{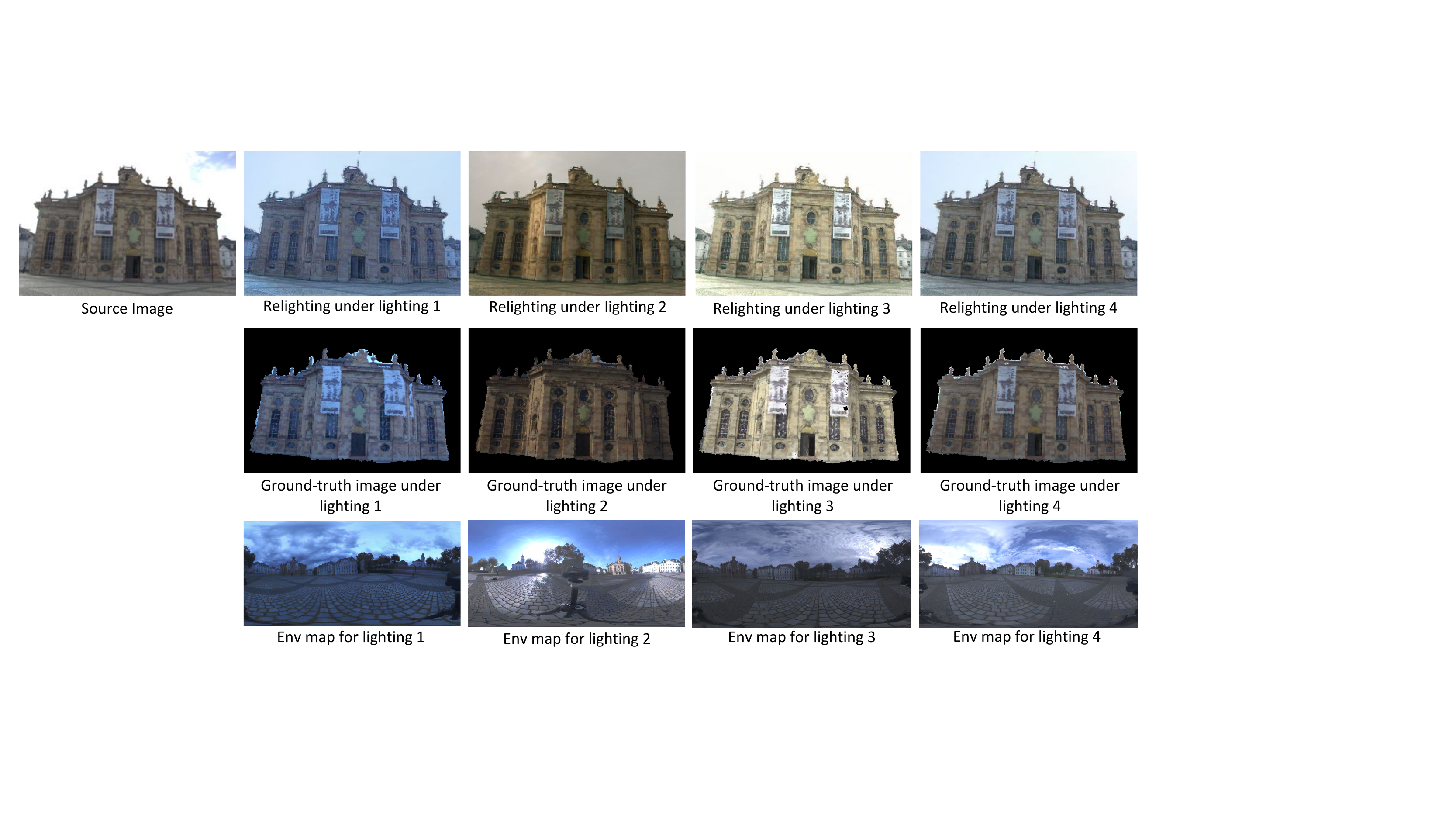}
	\caption{Relighting result on our new high-quality outdoor religting dataset. Note the plausible shading effects obtained by our method on the surfaces of the monument compared to the ground-truth.}
	\label{fig:dataset_result_1}
\end{figure}

\begin{figure}[!t]
    \centering
\begingroup
\setlength{\tabcolsep}{1pt}
\renewcommand{\arraystretch}{0.5}
    \resizebox{\textwidth}{!}{
\small{
\begin{tabular}{ccccccc}
\footnotesize{Input} & \footnotesize{Novel Illu1} & Our results & \cite{BarronTPAMI2015} & \footnotesize{Novel Illu2} & Our results & \cite{Yu_2019_CVPR}\\
\includegraphics[width=1.6cm]{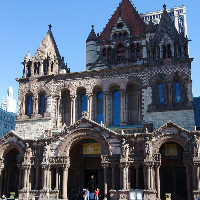}&
\includegraphics[width=1.6cm]{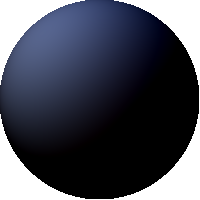}&
\includegraphics[width=1.6cm]{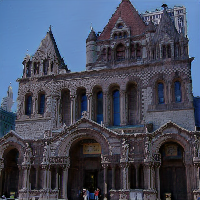}&
\includegraphics[width=1.6cm]{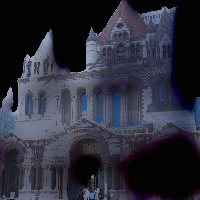}& 
\includegraphics[width=1.6cm]{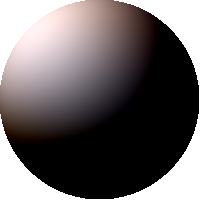}&
\includegraphics[width=1.6cm]{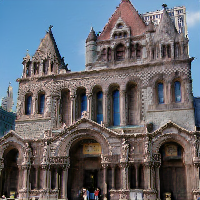}&
\includegraphics[width=1.6cm]{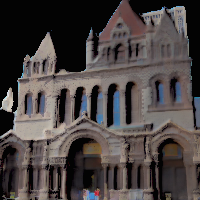}
\\

\includegraphics[width=1.6cm]{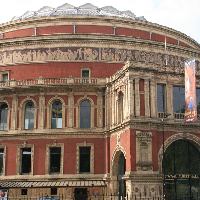}&
\includegraphics[width=1.6cm]{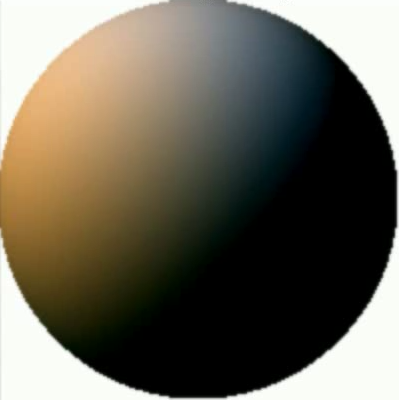}&
\includegraphics[width=1.6cm]{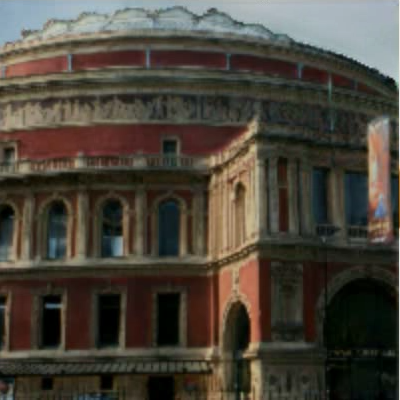}&
\includegraphics[width=1.6cm]{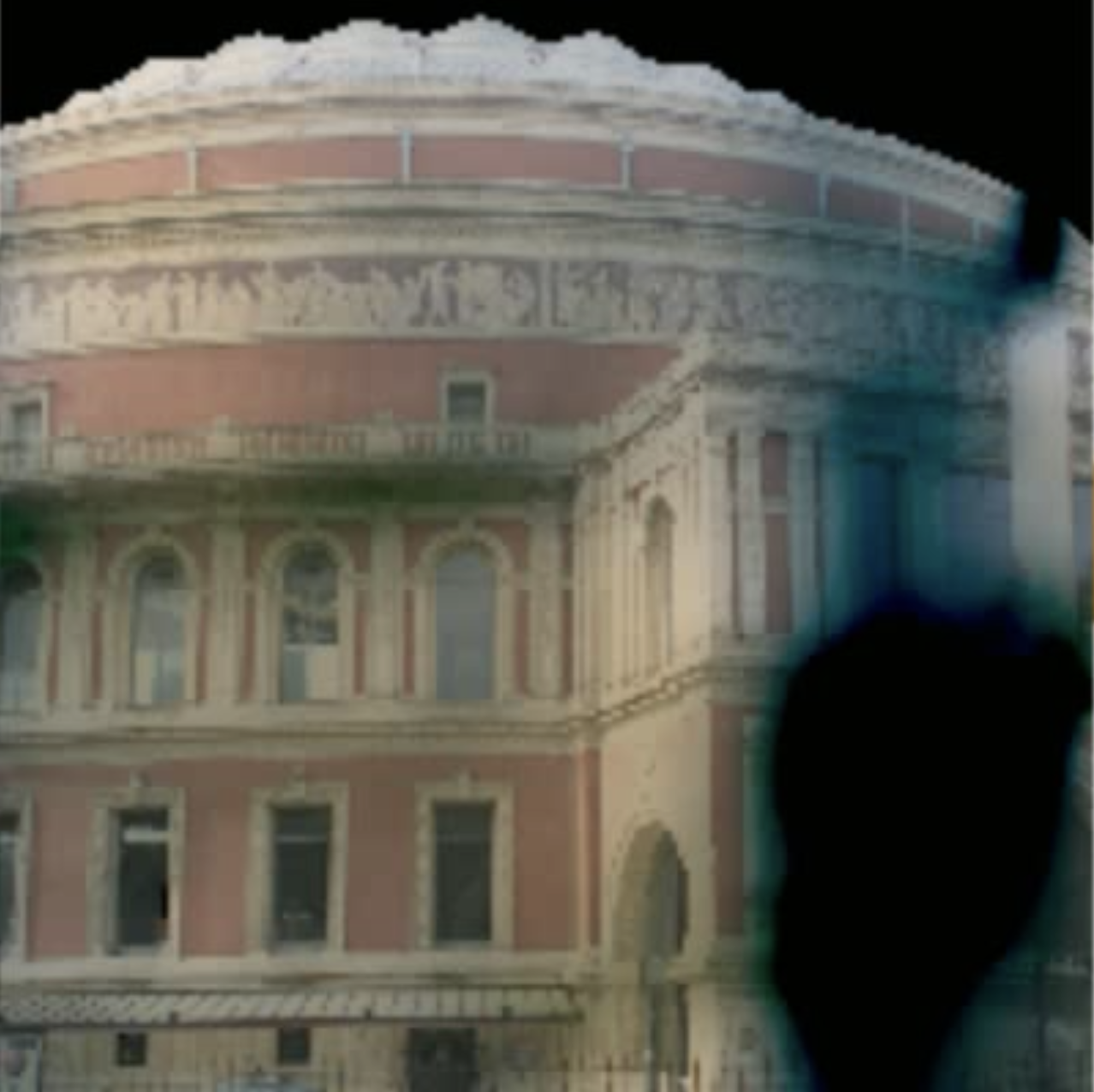}& 
\includegraphics[width=1.6cm]{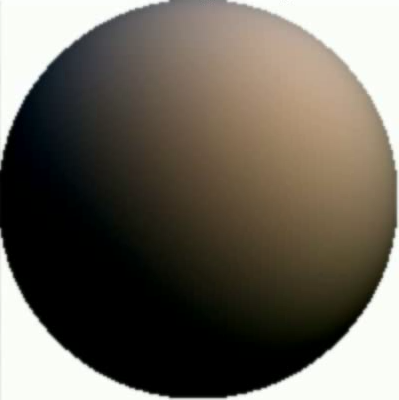}&
\includegraphics[width=1.6cm]{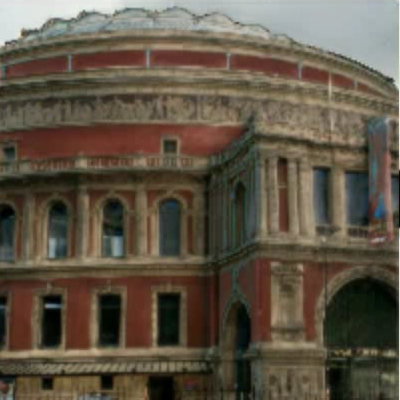}&
\includegraphics[width=1.6cm]{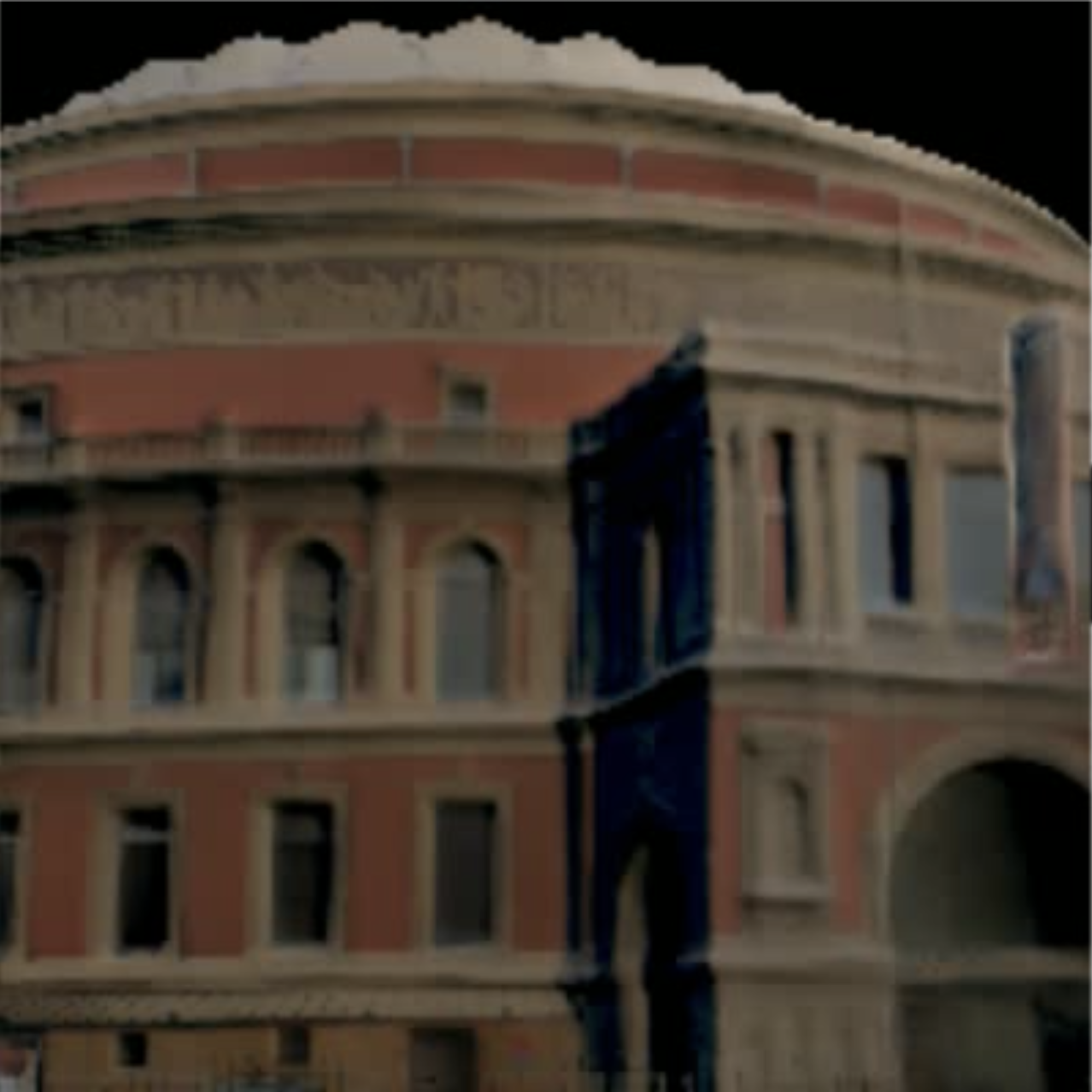}
\\

\includegraphics[width=1.6cm]{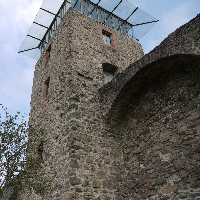}&
\includegraphics[width=1.6cm]{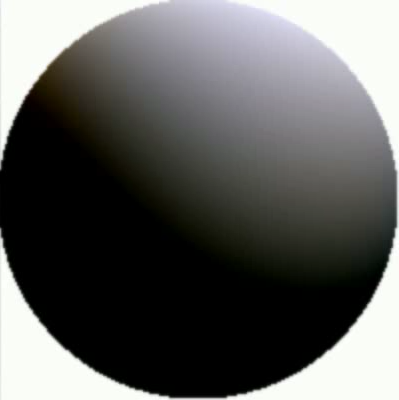}&
\includegraphics[width=1.6cm]{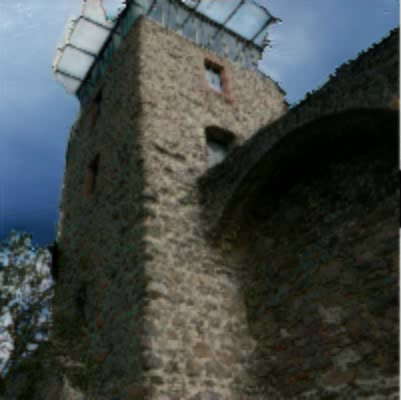}&
\includegraphics[width=1.6cm]{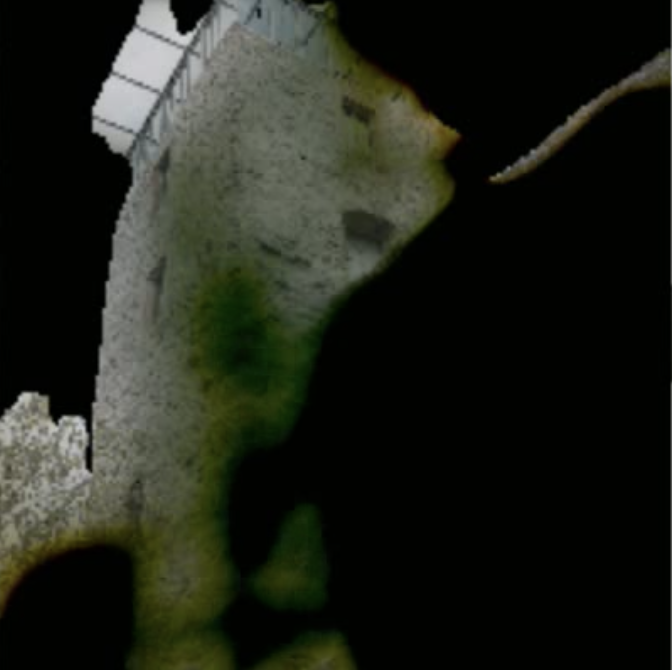}& 
\includegraphics[width=1.6cm]{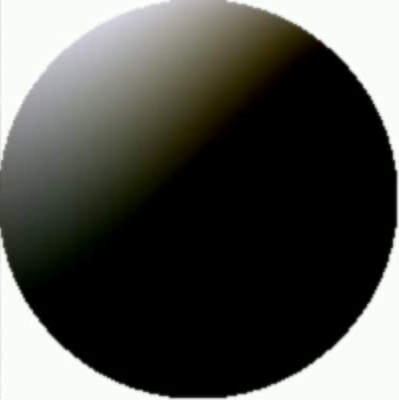}&
\includegraphics[width=1.6cm]{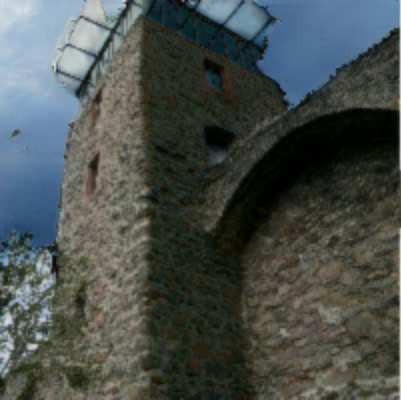}&
\includegraphics[width=1.6cm]{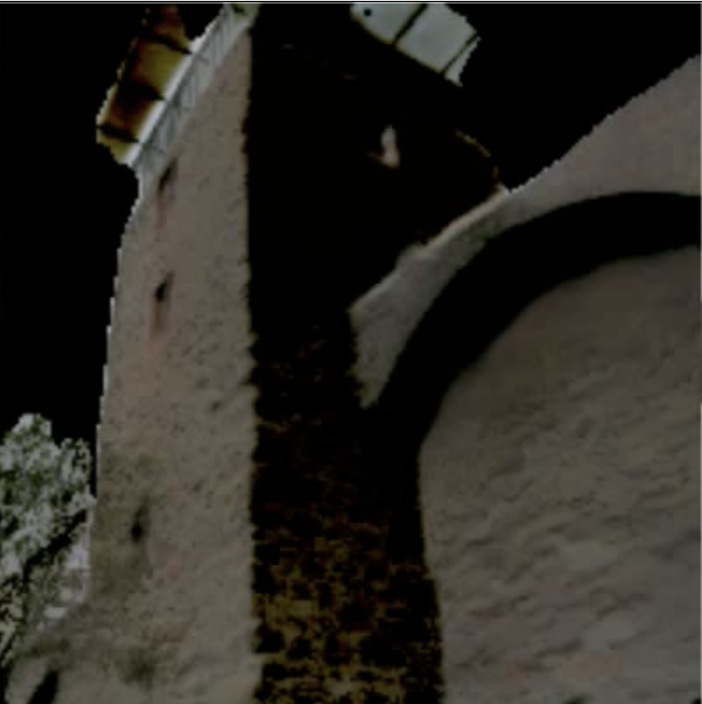}
\\

\end{tabular}
}}
\endgroup
    \caption{Relighting results from testing data. It shows the comparison between our methods with InverseRenderNet \cite{Yu_2019_CVPR} and SIRFS \cite{BarronTPAMI2015}.}
    \label{fig:test_results}
\end{figure}

\begin{figure}[!t]
    \centering
\begingroup
\setlength{\tabcolsep}{1pt}
\renewcommand{\arraystretch}{0.5}
    \resizebox{\textwidth}{!}{
\small{
\begin{tabular}{ccccccc}
\footnotesize{Source image} & \footnotesize{Target image} & \footnotesize{GT illumination} & Our results & \cite{Philip:2019:MRU:3306346.3323013} & \cite{BarronTPAMI2015} & \cite{Yu_2019_CVPR} \\ % 
\includegraphics[width=2.4cm]{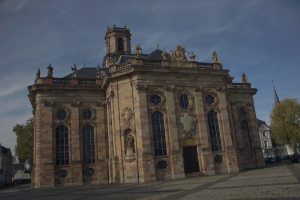}&
\includegraphics[width=2.4cm]{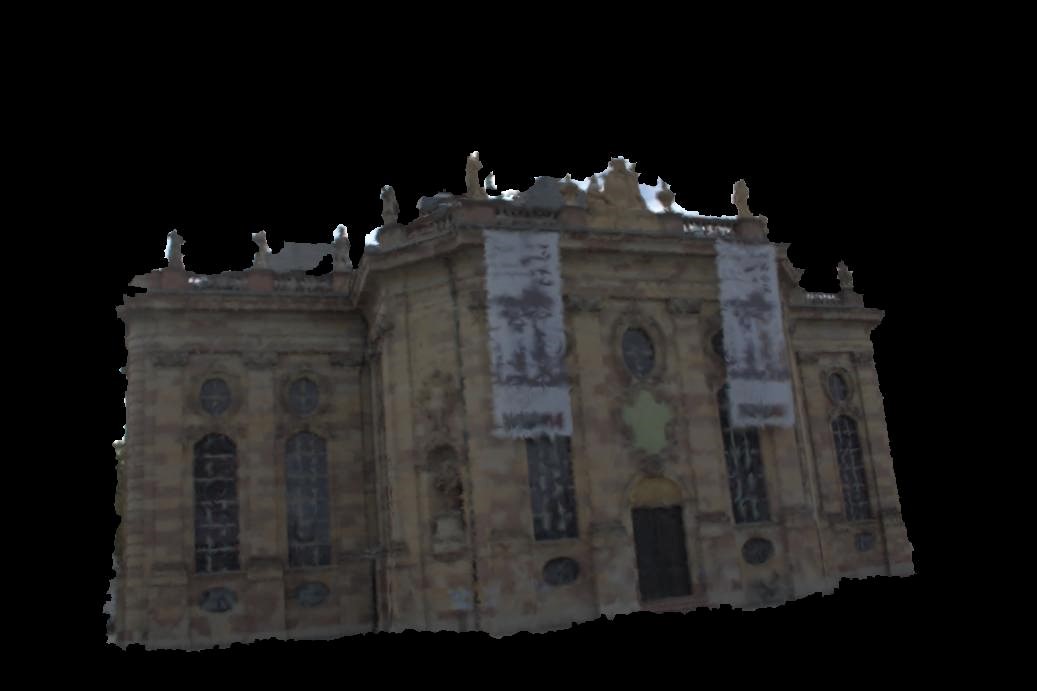}&
\includegraphics[width=2.4cm]{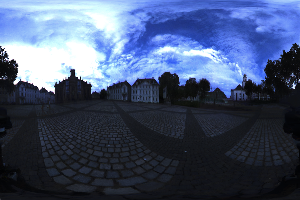}&
\includegraphics[width=2.4cm]{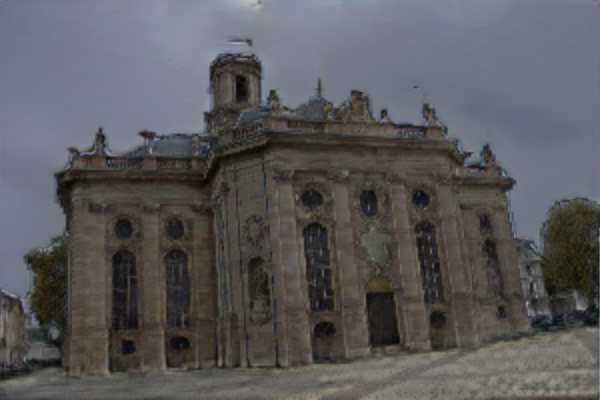}&
\includegraphics[width=2.4cm]{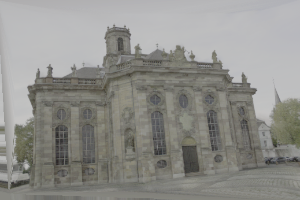}&
\includegraphics[width=2.4cm]{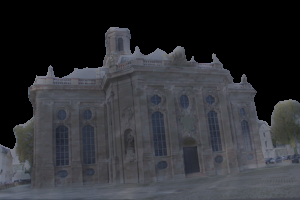}&
\includegraphics[width=2.4cm]{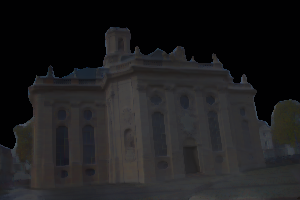}
\\
\includegraphics[width=2.4cm]{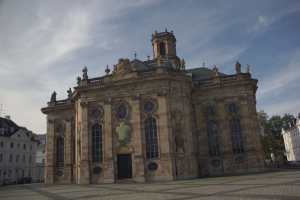}&
\includegraphics[width=2.4cm]{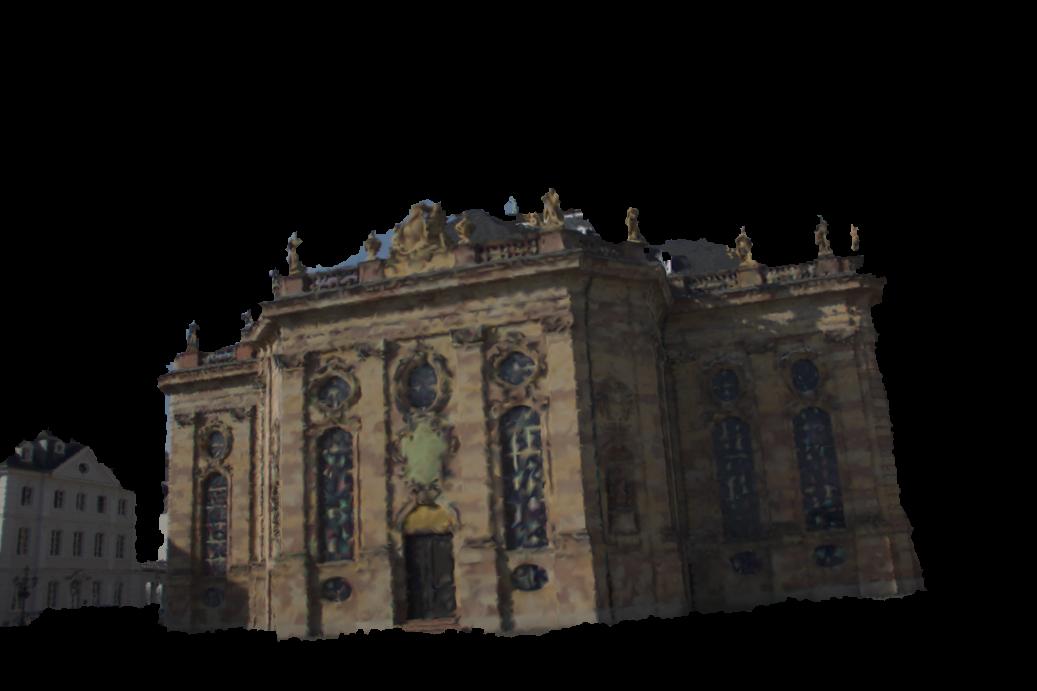}&
\includegraphics[width=2.4cm]{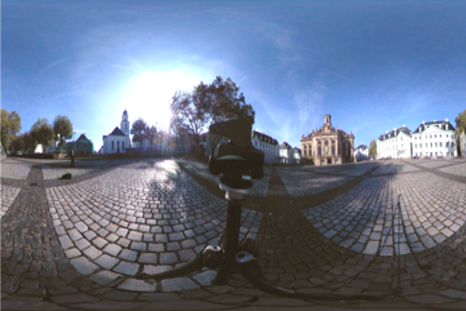}&
\includegraphics[width=2.4cm]{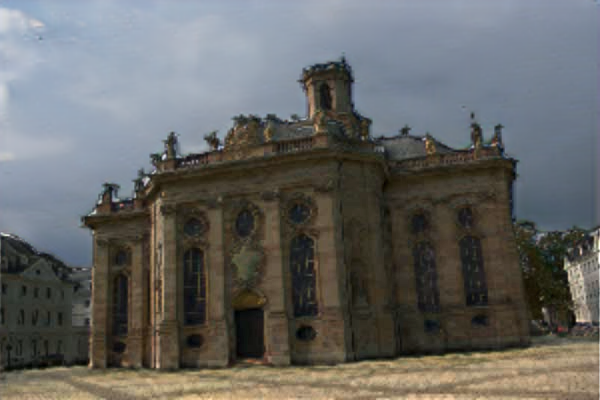}&
\includegraphics[width=2.4cm]{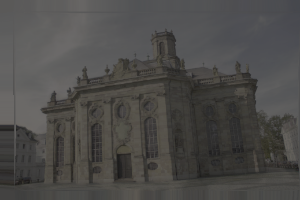}&
\includegraphics[width=2.4cm]{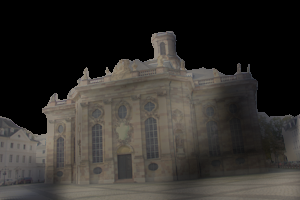}&
\includegraphics[width=2.4cm]{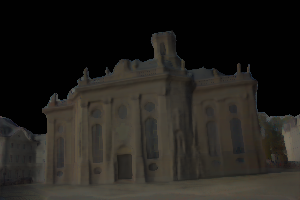}
\\

\end{tabular}
}}
\endgroup
    \caption{Relighting of benchmark dataset images and comparison with Philip \etal \cite{Philip:2019:MRU:3306346.3323013}, Yu and Smith \cite{Yu_2019_CVPR} and Barron and Malik \cite{BarronTPAMI2015}. }
    \label{fig:dataset_result_2}

\end{figure}

\begin{figure}[!t]
    \centering
\begingroup
\setlength{\tabcolsep}{1pt}
\renewcommand{\arraystretch}{0.5}
    \resizebox{\textwidth}{!}{
\small{
\begin{tabular}{ccccc}
\footnotesize{Source image} & \footnotesize{Target image} & Our results & \cite{BarronTPAMI2015} & \cite{Yu_2019_CVPR} \\ % 
\includegraphics[width=2.4cm]{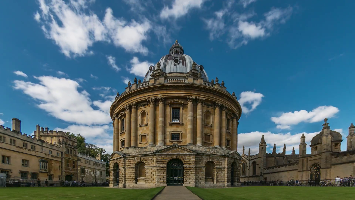}&
\includegraphics[width=2.4cm]{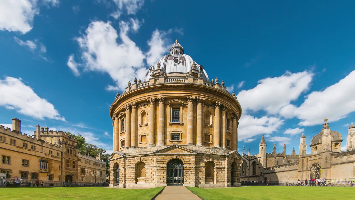}&
\includegraphics[width=2.4cm]{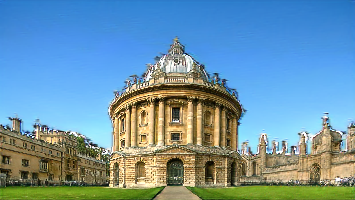}&
\includegraphics[width=2.4cm]{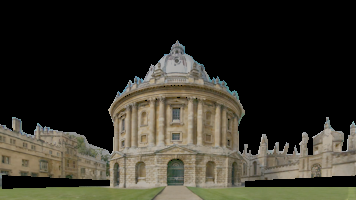}&
\includegraphics[width=2.4cm]{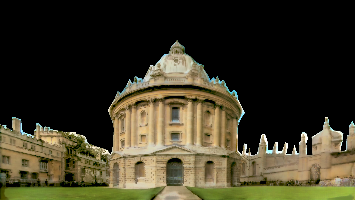}
\\
\includegraphics[width=2.4cm]{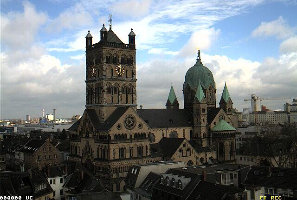}&
\includegraphics[width=2.4cm]{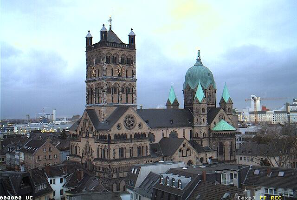}&
\includegraphics[width=2.4cm]{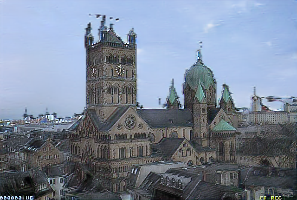}&
\includegraphics[width=2.4cm]{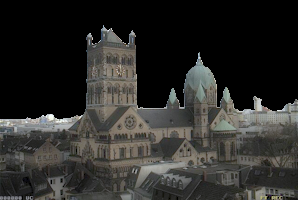}&
\includegraphics[width=2.4cm]{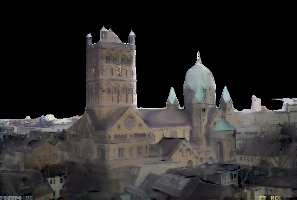}
\\

\end{tabular}
}}
\endgroup
    \caption{Relighting of BigTime images and comparison with Yu and Smith \cite{Yu_2019_CVPR} and Barron and Malik \cite{BarronTPAMI2015}. }
    \label{fig:bt}

\end{figure}

\mypara{On the Benchmarking Dataset} Our benchmarking dataset is used for qualitative evaluation of our method. We perform cross-relighting of the monument by taking an image for a particular lighting condition as input and performing relighting to another target light condition using as input the 2nd order spherical harmonic co-efficients of the ground-truth `aligned' environment light map. The results for such relighting is shown in Fig. \ref{fig:dataset_result_1}, where an image captured under a source lighting is relighted to several target lighting conditions, and Fig~\ref{fig:dataset_result_2}, where source image is relighted to target image under GT illumination. As can be seen, our method is able to generate relighting result that closely resembles the ground-truth images for each target lighting condition. Our method does a particularly good job of estimating plausible color-cast and shading across various surfaces of the monument including those with intricate geometry. 

\mypara{On test dataset} In Figure \ref{fig:test_results}, we show relighting results on our test split from MegaDepth data and comparison with other single-image relighting approaches. 
Our method results in realistic looking relighting results with shading and shadows that are very consistent with the target lighting condition, while maintaining the fine underlying reflectance details. We also generate sky regions which match the general colour tone of the relit structure. The method of Yu and Smith \cite{Yu_2019_CVPR} generates non-photorealistic images due to their simple Lambertian reflectance model. The method of Barron and Malik \cite{BarronTPAMI2015} struggles with the darker sides of the target lighting conditions because they cannot account for global illumination.

\mypara{On time-lapse dataset} We evaluate our neural rendering network on BigTime\cite{li2018learning} dataset, which contains approximately 200 time-lapse image sequences of indoor and outdoor scenes. 
For each time lapse sequence, we perform cross-rendering by relighting each frame with lighting estimates from all the other frames in the sequence.
The qualitative comparison between our method and other methods is shown in Figure \ref{fig:bt}.
It is evident that our method preserves the colour-cast and the brightness scale better, and is able to generate accurate relighting effects such as consistent shading and shadows.

\vspace{0em}
\subsection{Quantitative Evaluation}

We also perform quantitative evaluations on BigTime\cite{li2018learning} and our benchmarking data. 
To evaluate the relit results on BigTime\cite{li2018learning}, we use multiple error metrics computed between the relit result and corresponding real image. The quantitative comparison, averaged over 15 sequences, is shown in Table \ref{tab:bt}. It is shown that our network can generalise well to time-lapse image sequences. Our method has the best performance on $\ell_{1}$ error and the mean square error (mse) and is comparable to the method of Barron and Malik \cite{BarronTPAMI2015} on metrics measuring structural information like SSIM and DSSIM. 
Barron and Malik's \cite{BarronTPAMI2015} method seems to perform slightly better on these metrics because their method tends to baking albedo/reflectance details into shading. While this leads to preserving details in the output and better SSIM score (depending on how close the target and source lighting are), it is in general not a desirable quality (see Fig.~\ref{fig:test_results} for failure cases). 
This issue with their method is concealed when evaluating this dataset since the relighting is based on their estimated lighting.

Figure \ref{fig:dataset_result_2} shows example of the cross-relighting that we perform across all lighting conditions in the benchmark dataset. In order to get the ground-truth image for our relighting, we project all the camera images from a given target lighting condition onto the 3D geometry of the monument and average them. This is then re-projected to the camera viewpoint of the source image to obtain the ground-truth relit image. Although this leads to the loss of view-dependent effects, it still provides a plausible ground-truth image with accurate shadows and shading. Error metric is computed as $\ell_{1}$ error averaged over the reprojected pixels of the monument, see Table \ref{tab:my_label}. Our method generates plausible relighting results close the the ground-truth image and produces the least error in most cases, while the other techniques struggle to preserve the high-frequency details, the colour-cast and the shading variations. For the method of Philip \textit{et. al.} \cite{Philip:2019:MRU:3306346.3323013}, we were able to obtain cross-relighting results only in specific cases since their sun-lighting model cannot be applied to cloud or evening skies. Only in one case, their method was able to outperform ours quantitatively. Please note that their method uses the full multi-view dataset for relighting whereas our method relights a single image.

More results and ablation study can be found in supplementary document.

\begin{table}[!t]
    \centering
    \resizebox{.5\columnwidth}{!}{
    \begin{tabular}{c|c|c|c|c}
    \hline
     Method & $\ell_{1}$ & mse & SSIM & DSSIM \\
     \hline
     Proposed                               & {\bf 0.103} & {\bf 0.021} & 0.760 & 0.120\\
     \cite{Yu_2019_CVPR}                    & 0.117       & 0.26 & 0.722  & 0.139 \\
     \cite{BarronTPAMI2015}                 & 0.115       & 0.24 & {\bf 0.770} & {\bf 0.115}\\
     \hline
    \end{tabular}
    }
    \vspace{0.2cm}
    \caption{Quantitative evaluation on the BigTime \cite{li2018learning} time-lapse dataset. The error values are computed by averaging over 15 sequences.}
    \label{tab:bt}
\end{table}

\begin{table}[!t]
    \centering
    \resizebox{\columnwidth}{!}{
    \begin{tabular}{c|c|c|c|c|c|c|c|c|c|c|c|c}
    \hline
     \multirow{3}{*}{Method} & \multicolumn{12}{c}{Original lighting condition} \\
     \cline{2-13}
      & \multicolumn{2}{c|}{1} & \multicolumn{2}{c|}{2} & \multicolumn{2}{c|}{3} & \multicolumn{2}{c|}{4} & \multicolumn{2}{c|}{5} & \multicolumn{2}{c}{6} \\
      \cline{2-13}
      & $\ell_{1}$ & SSIM & $\ell_{1}$ & SSIM & $\ell_{1}$ & SSIM & $\ell_{1}$ & SSIM & $\ell_{1}$ & SSIM & $\ell_{1}$ & SSIM \\
     \hline
     Proposed                               & {\bf 0.077} & 0.871 & {\bf 0.078} & {\bf 0.850} & {\bf 0.074} & {\bf 0.876} & {\bf 0.075} & 0.872 & {\bf 0.076} & 0.842 & {\bf 0.073} & {\bf 0.839} \\
     \cite{Yu_2019_CVPR}                    & 0.082       & 0.824 & 0.085  & 0.780 & 0.087 & 0.791 & 0.083 & 0.818  & 0.079 & 0.819 & 0.077 & 0.810\\
     \cite{BarronTPAMI2015}                 & 0.083       & {\bf 0.879} & 0.097 & 0.826 & 0.091 & 0.852  & 0.080 & {\bf 0.883} & 0.086 & 0.840 & 0.098 & 0.814 \\
     \cite{Philip:2019:MRU:3306346.3323013} & &  & &  & &  & & &  0.095$^\dagger$ & {\bf 0.871} & 0.083$^\ddagger$ & 0.834 \\
     \hline
    \end{tabular}
    }
    \vspace{0.2cm}
    \caption{Mean $\ell_{1}$ colour error (lower is better) and SSIM index (higher is better) for relit images against cross projected ground-truth. Results are averaged across all images and all target lighting conditions. ($^\dagger$averaged over only target lighting condition 6 because the authors of method provided their results for only one target lighting condition.)($^\ddagger$averaged over only target lighting conditions 2 \& 5 for the same reason.)}
    \label{tab:my_label}
\end{table}

%% file: 7_discussion.tex
While our method generalizes well to various new scenes, it may be ill-posed for darker input images because sufficient information is not available due to limited photometric resolution of the camera sensor at lower light intensity levels to perform an accurate decomposition. Our method also struggles with  strong cast shadows. For similar reasons, a strong cast shadow in the input is a challenge for the inverse rendering network because it leads to non-linearity in the pixel-value vs. radiance curve which is difficult to recover. Conversely, generating strong cast shadows is also a challenge for the neural renderer. Generating such shadows involves simulating the physical ray-tracing process which requires a knowledge of the full 3D scene geometry. An interesting way of dealing with this would be to ensure that the rendering network is aware of such inaccuracies in the decomposition by training the entire pipeline end-to-end and make the network implicitly aware of the 3D scene geometry.

Our method, while capable of generating non-lambertian effects and thus relighting results with greater realism,
does not explicitly model them. This may sometimes lead to incorrect specularities that are not accurate reflections based on the position of the light source, the surface normals and the viewing direction. An explicit non-lambertian reflectance model and decomposition and a corresponding neural rendering pipeline would solve such an issue.

%% file: 8_conclusion.tex
We present a novel self-supervised single-image based relighting framework for outdoor scenes and an outdoor relighting benchmark dataset. This neural rendering framework based on self-supervision from casual photography can also be extended in the future to lighting augmentation tasks such as addition or removal of existing light sources in the scene, opening up interesting applications in augmented and virtual reality domain.